\documentclass[runningheads]{llncs}
\usepackage{graphicx}
\usepackage{comment}
\usepackage{amsmath,amssymb} 
\usepackage{color}
\usepackage{epsfig}
\usepackage{subfigure}
\usepackage[ruled]{algorithm2e}
\usepackage{multirow}
\usepackage{setspace}

\begin{document}
\pagestyle{headings}
\mainmatter
\def\ECCVSubNumber{4840}  

\title{SAMOT: Switcher-Aware Multi-Object Tracking and Still Another MOT Measure} 

\titlerunning{SAMOT}
%
\author{Weitao Feng\inst{1} \and
Zhihao Hu\inst{2} \and
Baopu Li\inst{3}\and
Weihao Gan\inst{4} \and
Wei Wu\inst{4} \and
Wanli Ouyang\inst{1}}
\authorrunning{Feng W. et al.}
%
\institute{The University of Sydney \and
Beihang University \and
Baidu USA\and Urban Computing Group, SenseTime}
\maketitle

\begin{abstract}
  Multi-Object Tracking (MOT) is a popular topic in computer vision. However, identity issue, i.e., an object is wrongly associated with another object of a different identity, still remains to be a challenging problem. To address it, switchers, i.e., confusing targets that may cause identity issues, should be focused. Based on this motivation, this paper proposes a novel switcher-aware framework for multi-object tracking, which consists of Spatial Conflict Graph model (SCG) and Switcher-Aware Association (SAA). 
  The SCG eliminates spatial switchers within one frame by building a conflict graph and working out the optimal subgraph. The SAA utilizes additional information from potential temporal switcher across frames, enabling more accurate data association. Besides, we propose a new MOT evaluation measure, Still Another IDF score (SAIDF), aiming to focus more on identity issues. This new measure may overcome some problems of the previous measures and provide a better insight for identity issues in MOT. Finally, the proposed framework is tested under both the traditional measures and the new measure we proposed. Extensive experiments show that our method achieves competitive results on all measures.
\end{abstract}


\section{Introduction}

\vspace{-2mm}
Multi-Object Tracking (MOT) is one of the fundamental problems in video analysis. 
Identity issues, i.e., an object is wrongly associated with another object of a different identity, are still challenging 
because of cluttered background, abrupt motion, and so on. To alleviate the identity issues, switchers, i.e., confusing targets that may cause identity issues, should be paid with great attention. In this paper, we try to handle switchers via two important steps: detection failures handling and data association.

\begin{figure}[t]
  \centering
  \includegraphics[width=4.1in]{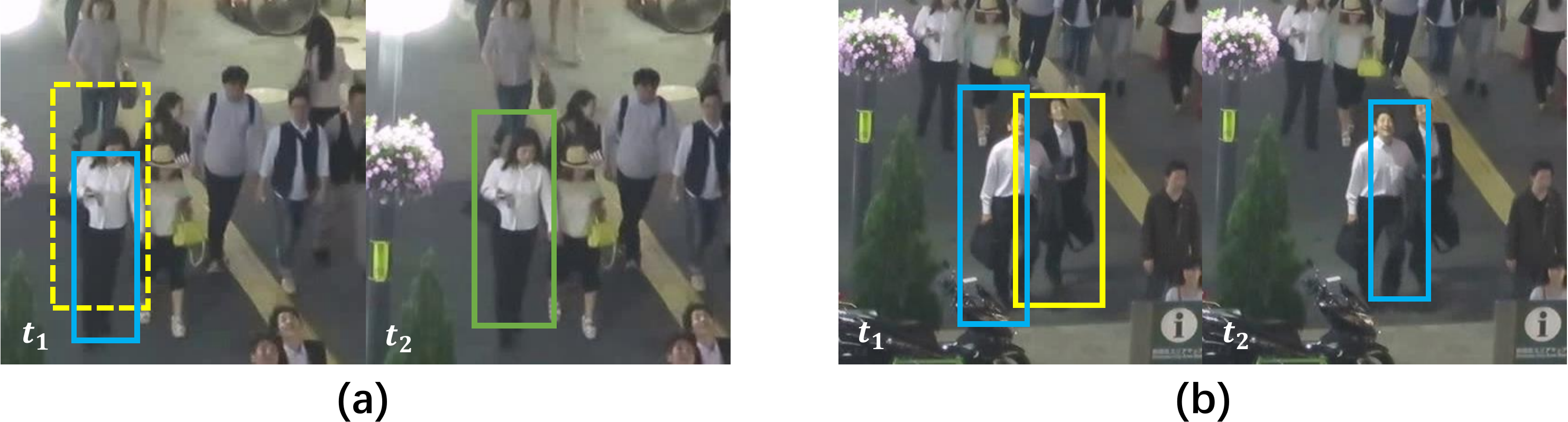}
  \vspace{-3mm}
  \caption{ (a) A hard case of false positive. The overlap of yellow dashed box and blue box in frame $t_1$ is less than 0.5 but they contain same target. Dashed box is false positive that confuses tracking and should be eliminated. (b) A hard case of association due to inaccurate localization. Comparison with potential temporal switcher in yellow leads to correct result. (Best viewed with online color version)}
  \label{fig:fig1}
\end{figure}

 False positives (FP) or inaccurate detections are one of the main reasons that lead to identity issues. We call these misleading detections within the same frame spatial switchers. When dealing with spatial switchers, most previous methods simply consider single detection and {ignore abundant spatial cues (e.g., occlusion relations) and temporal cues (e.g., consistency cross frames). If we utilize all these cues, it may be easier to find out false or inaccurate detections.} As shown in Fig.~\ref{fig:fig1}(a), after non-maximum suppression (NMS), both of the two boxes in frame $t_1$ still contain the same target but there is only one detection in frame $t_2$. The yellow dashed box is actually a false positive. If the yellow box is not eliminated, the tracker may wrongly associate this box to the green box in frame $t_2$, which causes the identity issue. By considering the spatial interaction between the yellow box and the blue box in frame $t_1$, we can consider the yellow dashed box as a spatial switcher that can be eliminated. 

 Association also plays a key role for MOT. 
 For association, most recent works \cite{zhu2018online,sadeghian2017tracking,Bergmann_2019_ICCV,son2017multi,Chu_2019_ICCV,long2018real} 
 use similarities to associate the coming detections and the tracked targets based on graph matching problem. 
 However, most of the previous approaches only focus on the similarity of the detection and one target, but ignore the effects from a confusing second target or more other targets in the previous frame. We call these additional targets across frames leading to identity issues temporal switchers. 
 Including information from potential temporal switchers tends to make the association easier. 
 As shown in Fig.~\ref{fig:fig1}(b), it is hard to tell whether the blue box in frame $t_1$ should be assigned to the blue box in frame $t_2$ without considering the potential temporal switcher due to inaccurate localization. 
 It is possible that features of the yellow box in frame $t_1$ and features of the blue box in frame $t_2$ are larger than the similarity between the two blue boxes in pairwise training.
 To avoid identity issue, we can add the yellow box in $t_1$ as the potential temporal switcher so that the blue box in $t_1$ will be correctly associated with the blue box in $t_2$.

 Motivated by the above observations, we propose an MOT framework that takes the spatial and temporal switchers into consideration. This switcher-aware framework consists of two parts, i.e., the Spatial Conflict Graph (SCG) and the Switcher-aware Association (SAA). SCG model is carefully designed to eliminate confusing spatial switchers, i.e., false positives and inaccurate detections. By working out the optimal subgraph of the SCG, we can filter out most spatial switchers. SAA includes switcher heat map and multiple appearance features of both the tracked targets and its potential temporal switcher, yielding more accurate association. These two parts are beneficial to each other and can reduce the identity issues effectively.

 When evaluating how serious the identity issues are, we find that previous metric (or measure) MOTA in CLEAR MOT Metric~\cite{bernardin2008evaluating} is hardly affected by the change of identity issues. We also find inefficiencies of the measure IDF1 score~\cite{ristani2016performance}. To better quantitatively validate the improvement for solving identity issues, we propose a new MOT evaluation measure, named SAIDF score. 
 The new measure may alleviate the problems of the CLEAR MOT Metric (include MOTA and IDS) and IDF1 score, it may also provide a better insight on identity issues. Analyses show that the new measure is sensitive to identity issues, 
 and it needs less calculation than IDF1. 

 In summary, our contributions are as below: 
\begin{itemize}
  \item \textbf{Spatial Conflict Graph (SCG) }:
  By utilizing the conflict graph and optimal sub-graphs, our proposed SCG can efficiently eliminate confusing spatial switchers in the same frame. 
  \item \textbf{Switcher-Aware Association (SAA)}:
  By combining switcher heat map and appearance features of both the tracked targets and its potential temporal switcher, more accurate data association can be achieved for MOT. 
  \item \textbf{SAIDF evaluation measure }:
    We propose a new SAIDF evaluation measure which focuses more on identity issues. This new measure may fix {the insensibility and high computational cost problems} of the previous measures and provide better insight about identity issues in MOT.
\end{itemize}
We comprehensively evaluate the proposed approaches on the challenging real-world dataset MOTChallenge and our method outperforms all the previous state-of-the-art(SOTA) methods in terms of all measures.
\section{Related Work}
\vspace{-2mm} 
\subsection{Approaches to Overcome Identity Issues}
\subsubsection{Detection Refinement.}
Detection refinement is taken as a pre-processing or post-processing step. For false negatives (FN), recent works like \cite{Chu_2019_ICCV,chu2017online,zhu2018online,xiang2015learning,yan2012track} use Single Object Trackers (SOT) of short term tracking as post-processing to reduce FN. Some SOT trackers like \cite{henriques2015high,bertinetto2016fully,danelljan2017eco,li2018high} have been applied in MOT. For FP or inaccurate detections, a few works like \cite{long2018real,Chu_2019_ICCV} take a binary classification network to refine the detection confidence and filter out wrong detections with a confidence threshold. 
However, most of them consider the problem with one single detection, ignoring the relationships between nearby detections.  In our work, we fully utilize the relationships between nearby detections to build a conflict graph, i.e., SCG, which is more effective to reduce confusing detections. {\cite{brendel2011multiobject} built similar graph model (Maximum Weight Independent Set) for MOT task, but our purpose is different thus the graph definition is different. We aim to eliminate confusing detection results while \cite{brendel2011multiobject} aims to work out the matching relationships.}  
\vspace{-4mm}

    \begin{figure}[t]
    \centering
    \includegraphics[width=3.8in]{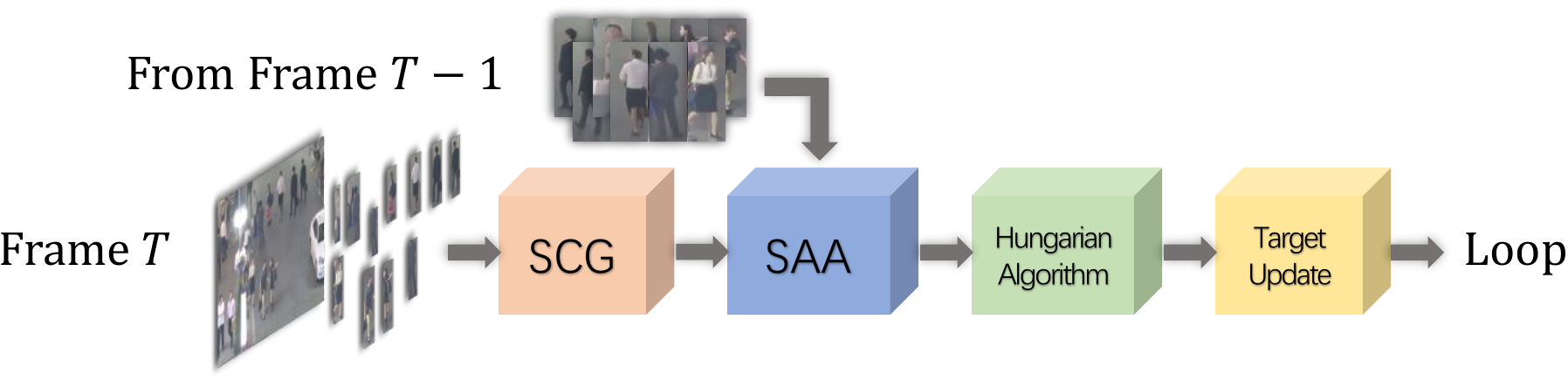}
     \vspace{-3mm} 
       \caption{Overall design of the proposed MOT framework. It consists of four major parts, the Spatial Conflict Graph (SCG) model for eliminating spatial switcher, the Switcher-Aware Association (SAA) for associating detection results, the Hungarian Algorithm (HA) and the target management.}
    \label{fig:arch}
    \end{figure}
    
\subsubsection{Data Association.}
The core problem for online data association is learning the matching scores on edges of the graph, also called metric learning. Some works like \cite{son2017multi,sadeghian2017tracking,milan2017online,bae2018confidence} emphasize novel features for obtaining better matching scores. 
Sadeghian \textit{et al.} \cite{sadeghian2017tracking} combine appearance, motion and interaction cues into a unified RNN network. Milan \textit{et al.} \cite{milan2017online} focus on the utilization of positions and motions of the targets. 
However, most methods mentioned above treat the metric learning as target-pair image classification without consideration of switchers (confusing identities). Our approach, i.e., SAA, takes both appearance and spatial information of the potential switcher to  handle the identity issues. 





\vspace{-3mm} 
\subsection{MOT Evaluation Measure}
The most famous CLEAR MOT Metric \cite{bernardin2008evaluating} utilizes MOTA , which averages false negatives, false positives and identity switches (IDS). However, in some applications of MOT, these mistakes are not equally evaluated. For long period video surveillance, identity issues, e.g., IDS, should be more emphasized. When evaluating a tracker with MOTA, even if we eliminate all IDS, the MOTA will not increase significantly. 

Besides MOTA, the IDS in CLEAR MOT Metric is not well treated.
First, it has a trivial solution (zero for empty output). Secondly, for example, if the tracker temporarily switches to another target and changes back to the correct target within a very short interval, it should be considered as a good action to fix the mistake, but it receives double penalty by IDS. 

When evaluating MOT in multi-camera task (MTMC), the typical measure is IDF1 \cite{ristani2016performance} score, which improves its sensitivity to identity issues across frames. However, IDF1 ignores most tracking results and considers the optimal largest segments. It is still not sensitive to some improvements on combining short identity segments. 

To alleviate the above problems, we design the SAIDF measure, improving the sensitivity to identity issues. Meanwhile, there is no trivial solution for it as for IDS. Unlike IDF1, our new measure considers all segments in the results and will be sensitive to any improvement. Besides, when computing, our method is much faster than IDF1. 

    \section{Switcher-Aware MOT Framework}
\vspace{-2mm} 
      An overview of the proposed framework is shown in Fig.~\ref{fig:arch}. For every frame $T$, the image and detection results first pass through the SCG to eliminate spatial switchers. Then the SAA predicts {more reliable} matching probabilities of the detection results from frame $T$ and the tracked targets from frame $T-1$. The SAA is followed by Hungarian Algorithm\cite{Munkres1957Algorithms} solving the matching relationships. The target management initializes the new targets and terminates the disappeared targets. This procedure will loop until no more frames are left.

      \begin{figure}
        \centering
        \includegraphics[width=4.0in]{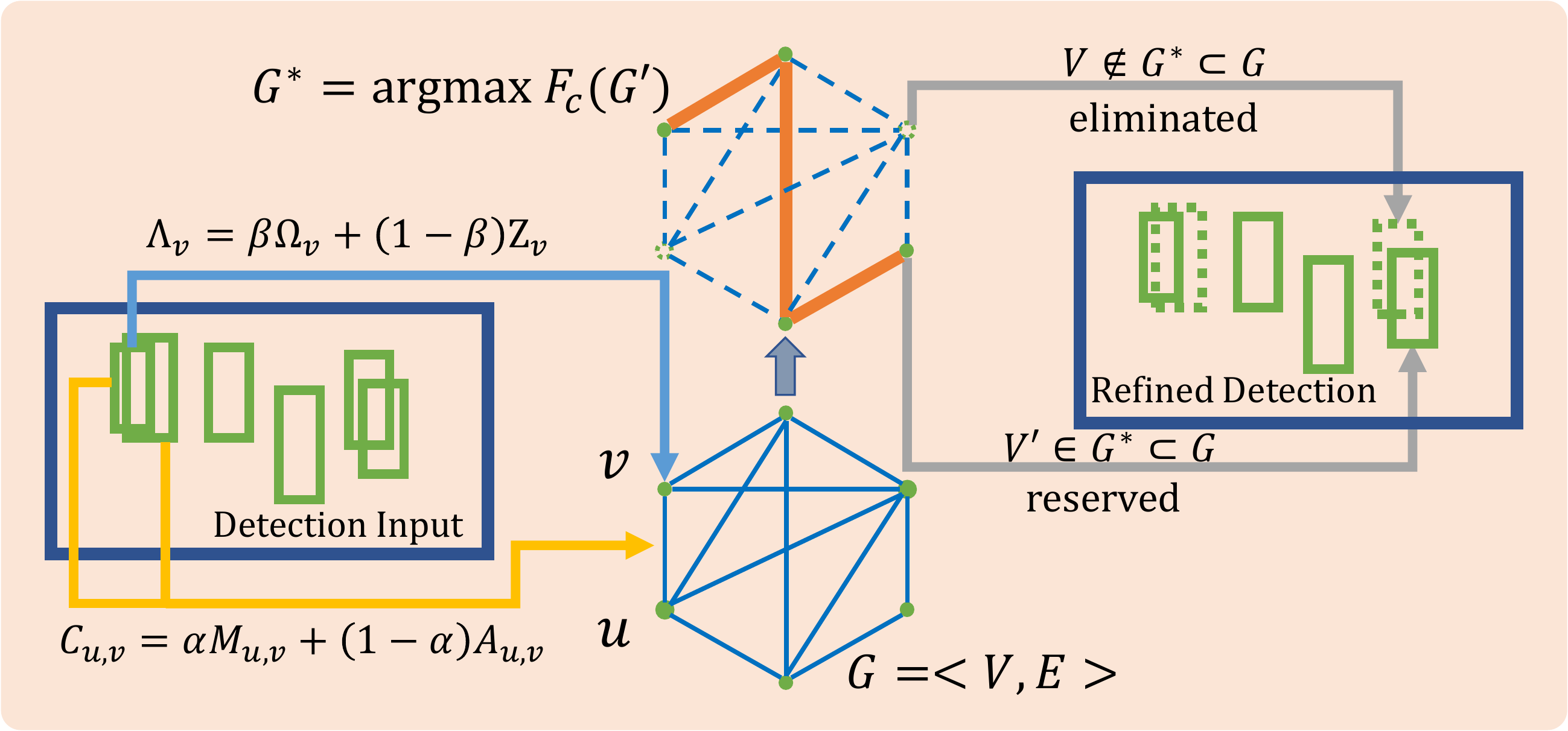}
         \vspace{-3mm} 
           \caption{The outline of SCG for eliminating spatial switcher. SCG is built with detections as nodes. Node and edge scores are calculated for the graph and then the optimal subgraph and corresponding detections are figured out.}
        \label{fig:SCG}
        \end{figure}    
      \subsection{Spatial Conflict Graph}
      \vspace{-1mm}      

    The pipeline of Spatial Conflict Graph (SCG) for eliminating spatial switcher is shown in Fig.~\ref{fig:SCG}, which can be considered as an optimal subgraph searching problem.
      In order to obtain the refined detection result $\mathcal{D}$, we build a conflict graph $G=<V, E;  \hat{\mathcal{D}}>$ on the raw detection results $\hat{\mathcal{D}}$ from external detector. Each node in the vertex set $V$ of $G$ stands for one detection result and each edge in the edge set $E$ means a conflict between two nodes. Then we figure out the optimal subgraph $G^*=<V^*, E^*; \mathcal{D}> \subset G$ from the following formulation: 
      \begin{equation}
        G^* = \mathop{\arg\max}_{G'=<E', V'>\subset G}{\sum_{u,v\in E'}{C_{u,v}}+\sum_{v\in V'}{\Lambda_v^2}},
      \end{equation}
      where $C_{u,v}\le 0$ is the conflict score between two nodes $u,v$, and $\Lambda_v\ge 0$ is the confidence score of node $v$. Therefore, the optimal subgraph can be considered as finding out the most confident nodes with less conflict between nodes.
      The corresponding detection results $\mathcal{D}$ in the optimal subgraph $G^*$ are the refined detection results. 

      \vspace{-2mm}  
        \bigskip\noindent\textbf{Conflict score.} For the graph $G$, the edge score that represents the conflict  $C_{u, v}$ for $u, v\in V$ can be calculated with:
        \begin{equation}
            C_{u,v}  = -\alpha M_{u, v} - (1-\alpha)A_{u,v},
        \label{eq:cuv}
       \end{equation}
        where $\alpha$ is a weight coefficient balancing the terms $M_{u,v}$ and $A_{u,v}$. 
        For the two features $F_a^u$ and $F_a^v$ generated by the appearance feature extractor in SAA,
         $A_{u,v}$ describes their cosine similarity as follows:
        \begin{equation}
            A_{u, v} = \frac{F_a^u \cdot F_a^v}{\|F_a^u\|\|F_a^v\|}.
        \end{equation}
        $M_{u,v}$ measures the occlusion confidence as follows:
        \begin{equation}
            M_{u,v} =\frac{\mathop{Int}(\mathop{Core}(u), \quad \mathop{Core}(v))}{  \mathop{Area}(\mathop{Core}(o))}, \textrm{where } o=
            \begin{cases} u,\  \mathrm{if}\ u\ \mathrm{ is\ the\ occludee},\\
            v, \  \mathrm{if}\ v\ \mathrm{ is\ the\ occludee},\\
            \end{cases}
         \label{eq:muv}    
        \end{equation}

        $Core(\cdot)$ in Eq.(\ref{eq:muv}) is the central region of the detection, which has $60\%$ of the width and height of the raw detection bounding box. $Int(\cdot, \cdot)$ measures the area of the intersected region for two input regions. $Area(\cdot)$ denotes the area of a region. $o$ denotes the occludee. The occlusion confidence $M_{u,v}$ describes the occlusion ratio of the core area of the occludee. In pedestrian tracking scenes, assume that $y_u^{bottom}$ denotes the bottom $y$-axis coordinate of $u$ and the same for $v$, when $y_u^{bottom} < y_v^{bottom}$, we assume that the detection result for node $u$ is the occludee occluded by node $v$ because $v$ is in front of $u$ from camera view, and vice versa. Note that the vanilla IoU is not able to describe such relationship between occluders and occludees in MOT scenes. Since $\mathop{Int}(\mathop{Core}(u), \quad \mathop{Core}(v)) \leq \mathop{Area}(\mathop{Core}(o))$, we have $M_{u,v}\leq 1$.
        
        \vspace{-2mm}  
        \bigskip\noindent\textbf{Confidence score.}
       For the graph $G$, the node score $\Lambda_v$ for the node $v\in V$ can be chosen as follows:
        \begin{equation}
        \Lambda_v = \beta \Omega_v+(1 - \beta)Z_v,
        \label{eq:lv}
        \end{equation}
        where $\beta$ is the coefficient balancing the terms $\Omega_v$ and $Z_v$. While $Z_v$, the detection score of $v$, is the classification score obtained from another binary classifier as some other works did \cite{long2018real,Xu_2019_ICCV}.
        $\Omega_v$, the consistency score for node $v$, is the max overlap between node $v$ and the detections in the previous frame defined as follows:
        \begin{equation}
            \Omega_v = \max_{d\in \mathcal{D}(t-1)}{\mathop{IoU}(d, v)}.
        \end{equation}

    The optimal subgraph can be solved in a very fast way, with the computational time required on par with the time required by Hungarian Algorithm on MOT17. Details for solving this problem is provided in the supplementary material. 
    

        \vspace{-4mm}
      \subsection{Switcher-Aware Association}
      \vspace{-1mm}      
       \begin{figure}[t]
        \centering
        \includegraphics[width=4.8in]{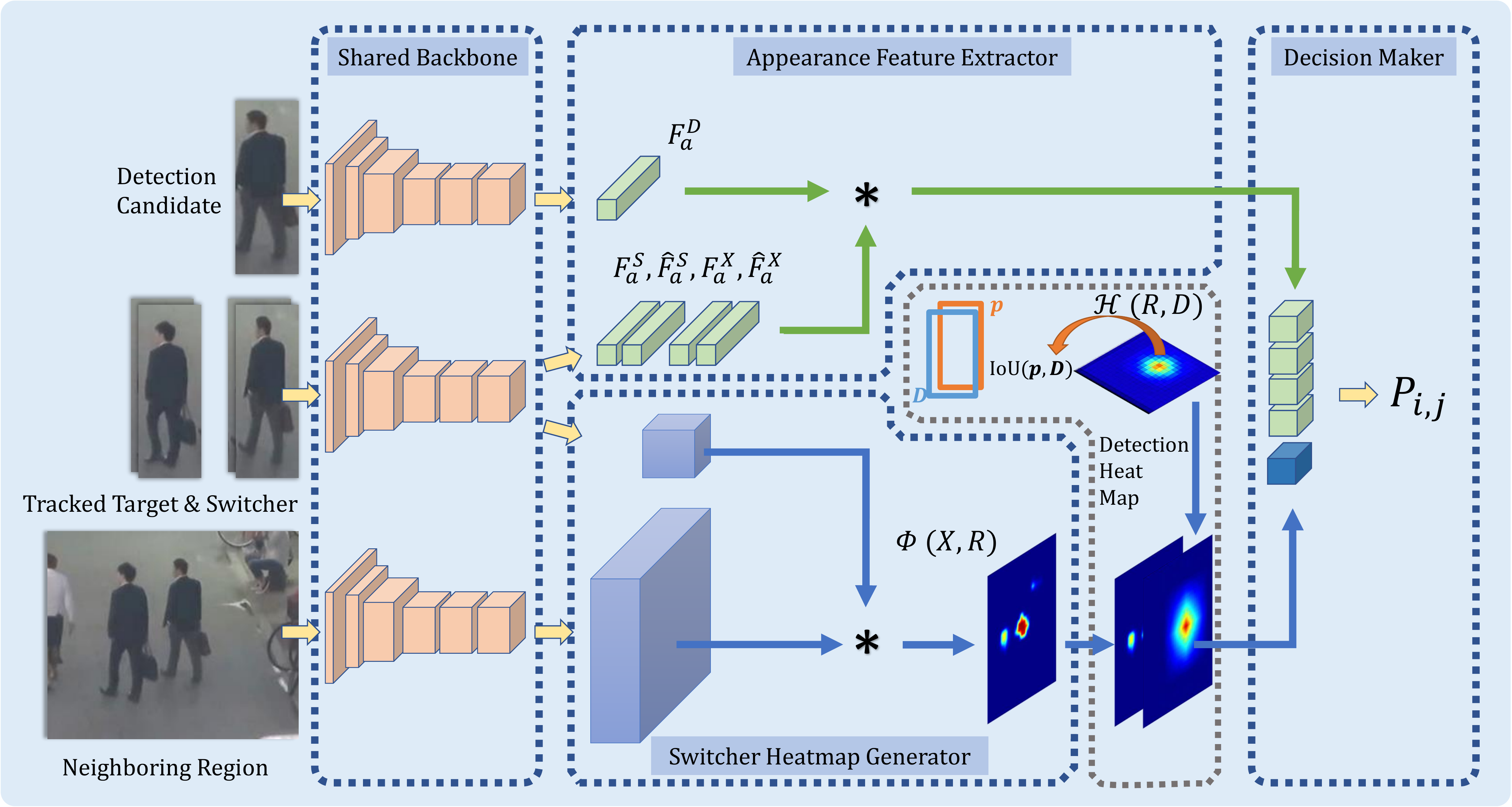}
         \vspace{-3mm} 
           \caption{Pipeline of the proposed switcher-aware association(SAA). There are four sub-nets in SAA: the shared backbone, the switcher heat-map generator, the appearance feature extractor and the decision-maker. The shared backbone yields general feature maps for the given images.The switcher heat-map generator takes feature maps from template of a target and the neighboring region as inputs. The CNN feature of the template is used as the kernel for convolution with the features of the neighboring region. 
              The appearance feature extractor takes several historical target images as inputs and uses the same weight-sharing backbone CNN and some extra header layers for extracting their historical appearance features. {The detection heat map indicates the relative position for the detection result (grey dashed area).}
              The decision-maker consists of some convolution layers and fully-connected (FC) layers. It combines the heat maps and the historical appearance features to make the final judgement.
           }
        \label{fig:SAA}
        \end{figure}

    The overview of SAA is shown in Fig.~\ref{fig:SAA} and the whole network can be trained end-to-end. SAA is composed of four parts: the shared backbone, the switcher heat-map generator, the appearance feature extractor, and the decision-maker. {We find out potential temporal switchers via switcher heat map and include them in data association.}
    For the backbone CNN, we use a modified ResNet-18 \cite{he2016deep} as the weight-sharing backbone of the appearance feature extractor and the switcher heat-map generator. More details about it can be found in supplementary file. The other three parts are introduced here one by one.

    \vspace{-2mm} 
      \bigskip\noindent\textbf{Switcher heat-map generation.} The feature maps from block2, block3 and block4 of the backbone are denoted by $F_2$, $F_3$ and $F_4$, respectively. The feature maps from the template of target $X$ are described as $F_2^X$, $F_3^X$ and $F_4^X$, and the feature maps from the neighboring region $R$ are represented by $F_2^R$, $F_3^R$ and $F_4^R$.
      Then the output of switcher heat-map generator can be described with the following equation:
       \begin{equation}
          \Phi(X, R) = \mathrm{sigmoid}(\frac{1}{3}\sum_{i=2,3,4}{F_i^X * F_i^R}),
       \end{equation} where $F_i^X * F_i^R$ denotes the convolution between the template $F_i^X$ and all possible locations in $F_i^R$, which is used for obtaining the similarity between the template and all possible locations in the neighboring region. The switcher heat map is supervised by binary cross entropy loss on map $L_{h}$ like \cite{li2018siamrpn++}.
       

          \vspace{-2mm}  
      \bigskip\noindent\textbf{Find the potential switcher from the switcher heat-map generator. } We identify the potential temporal switcher from the switcher heat map using a modified Non-Maximum Suppression (NMS) algorithm. We first find out the second-best position {and its corresponding bounding box (same size as the template at the position)} satisfying two conditions: 1) its {bounding box overlap with the one of the first largest-response position} is smaller than 0.5; 2) it has the largest response score under condition 1). If the response score of the second-best position is larger than a threshold of 0.5, then we use the target that is nearest to the second-best position in previous frame as the potential switcher; otherwise, {it is considered that no potential switcher is found and} we will use the IoU of target and detection result as the matching probability instead for efficiency. After the potential switcher is identified, its appearance features will be extracted 
      and used as part of inputs to the final decision-maker.
    
    
              \vspace{-2mm}  
      \bigskip\noindent\textbf{Appearance Feature Extractor.}
      The feature maps from block4 of the backbone are then passed through a group of header layers.
      After the header layers, the appearance feature of a target is calculated. {For a tracked target $X$, $F_a^X$ is the appearance feature of the target, $F_a^S$ is the appearance feature of the potential switcher and $F_a^D$ is the appearance feature of the detection result. }
      {To supervise the feature learning, we add another binary cross entropy (BCE) loss 
      \begin{equation}
          L_{a} = \frac{1}{2}L_{BCE}(\eta F_a^X * F_a^D, y_x) + \frac{1}{2}L_{BCE}(\eta F_a^X * F_a^S, y_s) 
      \end{equation}
      for the appearance feature extractor. Here $\eta$ is the rescaling weight, and $y_x$ is the corresponding label of whether the target and the detection are identical. $y_s$ have similar meaning for the target and potential switcher. }

      \label{sec:APPR}
       Due to frequent interactions of the targets and occlusions, we find that in some critical frames the bounding boxes are inaccurate and therefore the corresponding appearance features may not be reliable. To handle these cases, we include extra appearance features as part of inputs of the decision-maker. $\hat{F_a^X}$ and $\hat{F_a^S}$ respectively to denote the saved appearance features of the target and potential switcher from the history, i.e., in the frames before $T-1$. 

              \vspace{-1mm}  
       \bigskip\noindent\textbf{Final Decision-Maker.} 
       {To embed position of the target and detection result $D$, we generate another detection heat map on the same scale with the switcher heat map and include it as part of the input of decision-maker. The detection heat map $\mathcal{H}(R, D): \mathbb{D} \rightarrow \mathbb{R}$ is defined as follow:}
      \begin{equation}
          \mathcal{H}(R, D)[\mathbf{p}] = \mathop{IoU}(\mathbf{p}, D),
      \end{equation}
      where $\mathbf{p} \in \mathbb{D}$ denotes the position on the map.
      As shown in Fig. \ref{fig:SAA} (grey dashed area), the detection heat map indicates the relative position of the detection result to the target.

       To make the final decision whether the relationship of the pair of targets should be considered matched or separated, the final decision-maker combines all cues generated in previous steps. As such, the input set can be denoted by
       \begin{equation}
           \Gamma = \{\Phi(X, R), \mathcal{H}(R, D), F_a^X, \hat{F_a^X}, F_a^S, \hat{F_a^S}, F_a^D\},
       \end{equation} {where $\Phi(X, R)$ is the switcher heat map, $\mathcal{H}(R, D)$ is the detection heat map, $F_a^X$, $F_a^S$, $F_a^D$, 
       $\hat{F_a^X}$ and $\hat{F_a^S}$ are the appearance features introduced before. }
       

      For the features $F^D_a$ of a detection candidate, we calculate its cosine similarity to the tracked target feature $F^X_a$, the historical feature $\hat{F^X_a}$, and similar for the potential switcher. 
      For heat maps, we concatenate the switcher heat map and the detection heat map channel-wisely and add two convolution layers with batch normalization layers after them, generating another embedded feature. We concatenate all these similarities and the embedded features of heat map, and then feed them to  fully-connected layers with ReLU\cite{nair2010rectified}, as shown in Fig.~\ref{fig:SAA}. Finally, the output is the matching probability of the detection candidate and the target, which will be used in the data association procedure.
      We add the third BCE loss $L_d$ for the decision-maker.
      \subsection{Hungarian Algorithm and Target Management}
    \vspace{-2mm} 
    

       Following most previous online MOT methods \cite{lee2018learning,bae2014robust,sanchez2016online,yu2016poi,wojke2017simple,bewley2016simple}, we run Hungarian Algorithm to find out the matching relationships maximizing the sum of matching probabilities generated by SAA.
       For any isolated detection result after Hungarian Algorithm, we add them as new identities. We use a confidence score to decide when to drop the lost targets. The score is set to the detection score when a new target is generated or a matched target is updated. For a tracked object, 
       if there is no edge connecting it when running the Hungarian Algorithm, then the target is considered as temporarily lost. In this case, we decay its confidence score by 0.95. If the decayed confidence score is still larger than a threshold, we will keep it, otherwise it will be dropped. For a matched target, its location will be updated as the detection result. While for an isolated target, its location will be renewed as the largest response location in its switcher heat map if the confidence score is larger than a threshold of 0.9, which is regarded as a reliable prediction of a single object tracker.
 
    \section{Still Another MOT Measure}
    \vspace{-2mm} 
    
    \subsection{Analysis on Existing MOT Measures}
    
    \subsubsection{Analysis on MOTA.} According to \cite{bernardin2008evaluating},
       \begin{equation}
           \mathrm{MOTA} = 1 - \frac{\sum_t{(m_t+{fp}_t+{mme}_t)}}{\sum_tg_t}.
       \end{equation}
       where $m_t$ is missed target, or false negatives (FN), in frame $t$, ${fp}_t$ is the false positives (FP) in frame $t$, ${mme}_t$ is the mismatched error (also called IDSwitch, IDS) in frame $t$. In challenging datasets like MOT17, published high ranking trackers (MOTA $\ge$ 48\%) have an average IDS, FP, FN of 3116.4, 22565.4 and 248712.8, which occupy 0.552\%, 4.000\% and 44.080\% of MOTA. Even if we eliminate all IDS, the MOTA will only increase no more than 0.552\%, which is far less than the improvement room of FP and FN. It shows the insensibility of MOTA to identity issues. In other words, the FN number can greatly influence MOTA. However, eliminating FN is more relative to the external detector or single object tracking.

    \vspace{-4mm} 
    \subsubsection{Analysis on IDS.} 
\begin{figure}[htbp]
        \centering
        \subfigure[]{
        \begin{minipage}[c]{0.46\textwidth}        \label{fig:IDS}
        \centering
        \includegraphics[width=2.4in]{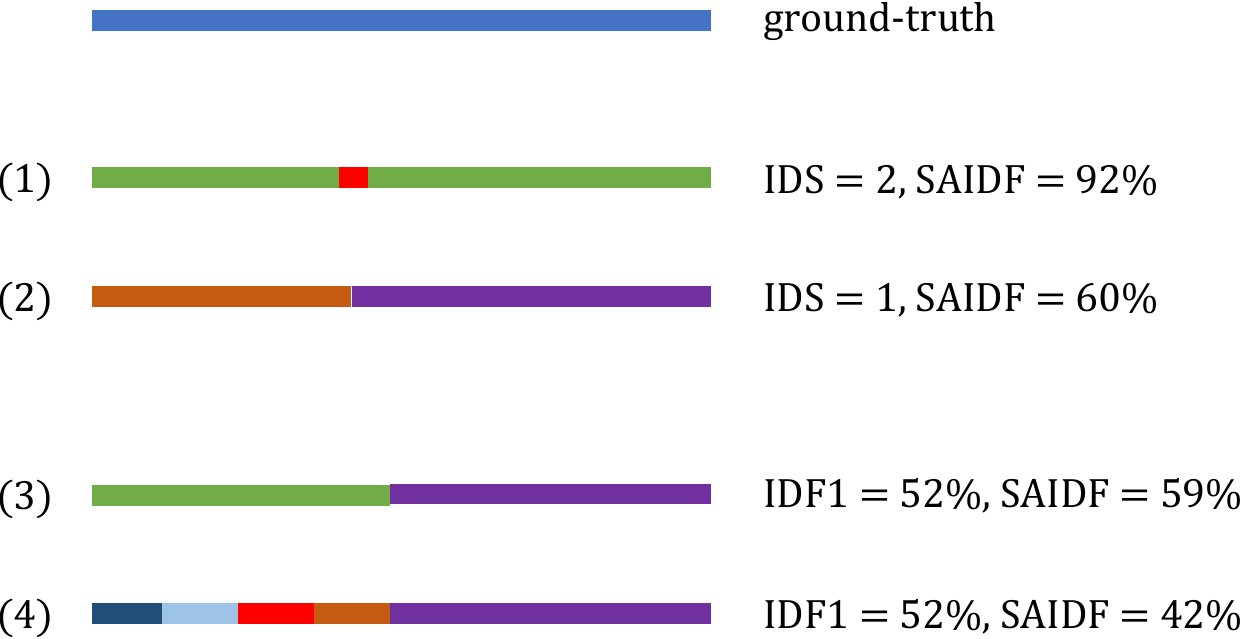}
        \end{minipage}
        }
        \subfigure[]{
        \begin{minipage}[c]{0.46\textwidth}        \label{fig:SEN}
        \centering
        \includegraphics[width=1.95in]{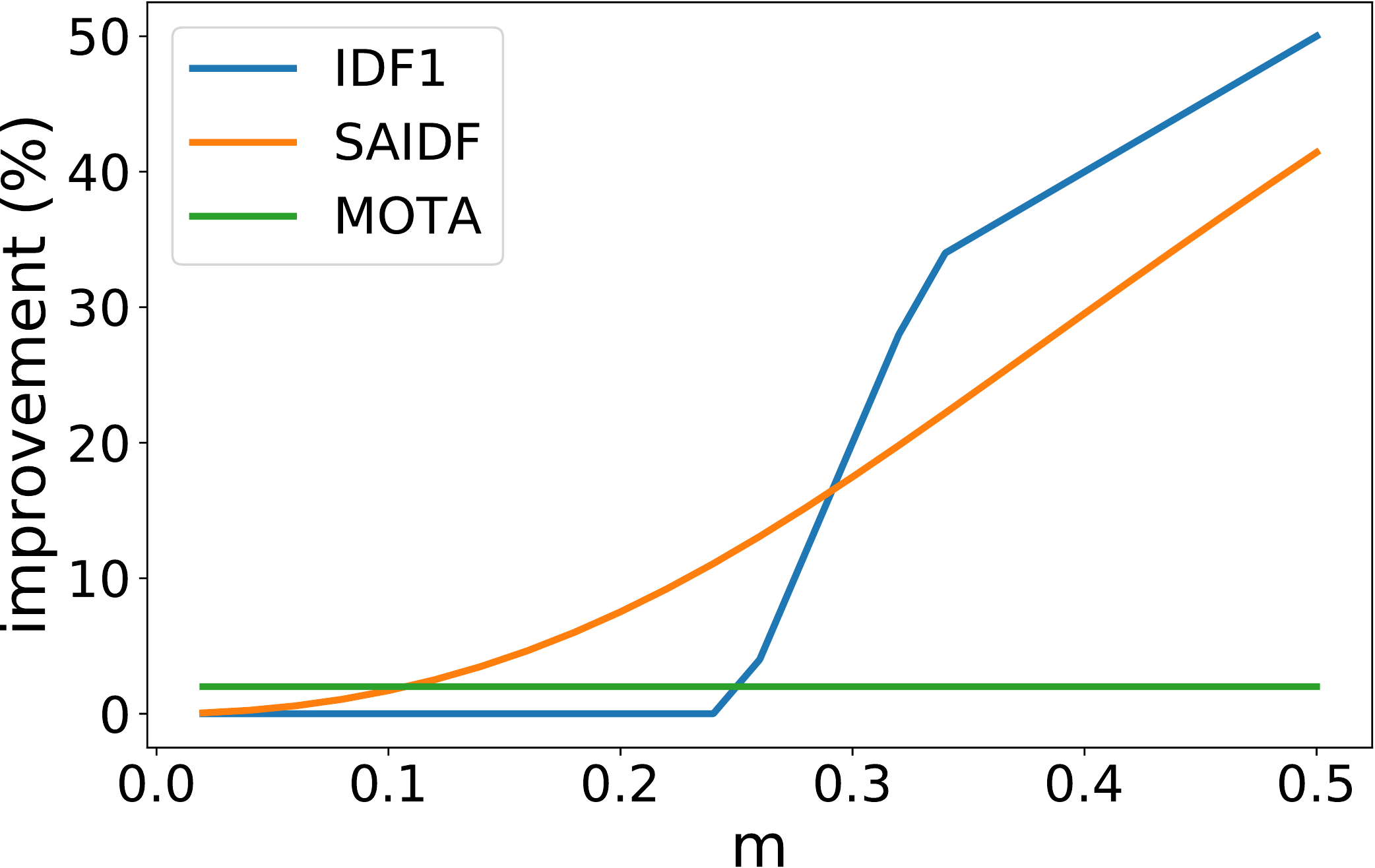}
        \end{minipage}
        }
          \vspace{-3mm} 
           \caption{(a): IDS, SAIDF, IDF1 and SAIDF for 4 different tracking results. Different colors stand for different hypotheses. (1) is a better tracking result than (2) but has higher IDS, (3) is a better tracking result than (4) but has same IDF1. The proposed SAIDF fixes all these drawbacks. (b): Analysis on sensitivity of MOTA, IDF1 and SAIDF. $m$ is the length ratio of the segments involved in solving identity issues.}
        \end{figure}

Identity issue refers to an object wrongly associated with another object of a different identity. Although IDS can roughly account for the degree of identity issues that happen when processing the video, it is not discriminative in some cases. 
    As shown in Fig. \ref{fig:IDS}, the tracking result (1) in Fig. \ref{fig:IDS} is a better tracking result than the result of (2), because the tracking result (1) is correct except for a very small time duration denoted by the red segment while the tracking result (2) has 40\% of the segment being incorrect. But (1) gets two IDS while (2) has only one. Besides, it is hard to evaluate identity issues only based on IDS. A trivial solution  outputting empty tracking result would have zero IDS. Therefore, IDS may not be a proper measure responsible for the overall tracking performance.


    \subsubsection{Analysis on IDF1.}   
    IDF1 overcomes the problem of IDS, but the IDF1 ignores all matched segments except the largest one. {Assume that a target is hard to track in the first half of the tracklet but becomes easy to track later. As shown in Fig.~\ref{fig:IDS}, the purple segments in tracking results (3) and (4) correspond to the easy case, while the green segment in tracking result (3) corresponds to the hard case. In this situation, an algorithm good at such hard case and producing the tracking result (3)  
may not get any improvement on the IDF1 when compared with the algorithm producing the tracking result (4) in Fig.~\ref{fig:IDS}. Such cases may frequently appear in the real world application, such as vehicles and pedestrians hard to be tracked initially but then easy to be tracked when waiting for traffic lights. Another issue for IDF1 is the speed of calculation. Due to global optimal matching, the calculation is very slow in longer and denser video scenes.}
   \vspace{-4mm} 
    \subsection{Definition of SAIDF Measure}
\label{sec:SAIDF}
 \vspace{-1mm} 
 To address the problems above, we propose Still Another IDF (SAIDF) score. In order to illustrate our proposed SAIDF, we first briefly introduce IDR and IDP for IDF1.
{With ground-truth trajectory set $\mathcal{O}$ and predicted trajectory/hypotheses set $\mathcal{Q}$, the IDR and IDP to calculate IDF1 can be rewritten as 
\begin{equation}
\label{eq:IDR_IDP_calculation}
\text{IDR} = \sum_{o\in \mathcal{O}}{W_o\cdot \eta_r(o)}, \ \ \text{IDP} = \sum_{q\in \mathcal{Q}}{W_q\cdot \eta_p(q)},
\end{equation}
where $W_o=\frac{\mathrm{len}(o)}{\sum_{o'\in \mathcal{O}}{\mathrm{len}{(o')}}}$ and $W_q=\frac{\mathrm{len}(q)}{\sum_{q'\in \mathcal{Q}}{\mathrm{len}{(q')}}}$ are the weights of the ground-truth trajectory and the predicted hypothesis, respectively.  $\mathrm{len}(\cdot)$ calculates the number of frames of a trajectory or a hypothesis. $\eta_r(o)$  and $\eta_p(q)$  are obtained from a global bi-partite matching, which cause the mentioned problems of IDF1.}

 We address the problems of IDF1 by defining 
    still another IDR (SAIDR) and still another IDP (SAIDP) as follows:
    \begin{equation}{
        \mathrm{SAIDR} = \sum_{o\in\mathcal{O}}{W_o\cdot\sqrt{\sum_{q\in\mathcal{Q}}{\lambda(o, q)^2}}}, \ \         \mathrm{SAIDP} = \sum_{q\in\mathcal{Q}}{W_q\cdot\sqrt{\sum_{o\in\mathcal{O}}{\lambda(q, o)^2}}},}
    \end{equation}
     where $W_o$ and $W_q$ are the same as those in ~Eq.(\ref{eq:IDR_IDP_calculation}), $\lambda(o,q)$ is the matched length in time over the union of a true trajectory and a hypothesis in time (similar to IoU). Denoting  $\mathcal{M}(o, q)$ as number of matching box pairs for $o$ and $q$ and $Intrs(o, q)$ as the number of frames that $o$ and $q$ appear concurrently, we may have

\begin{equation}
    \lambda(o, q) = \frac{\mathcal{M}(o, q)}{\mathrm{len}(o) + \mathrm{len}(q) - Intrs(o, q)}. 
\end{equation}
We also follow the truth-to-result match procedure in \cite{ristani2016performance} to produce matched segments, so the final SAIDF is defined as below:
    
    \begin{equation}
        \mathrm{SAIDF} = \frac{2 \cdot \mathrm{SAIDR} \cdot \mathrm{SAIDP}}{\mathrm{SAIDP} + \mathrm{SAIDR}}.
    \end{equation}
    \vspace{-4mm} 
    \subsection{Analysis on SAIDF Measure}
    \vspace{-1mm} 

    \subsubsection{Trivial Case.} When the hypotheses perfectly match the true trajectories, SAIDR and SAIDP are 100\% and SAIDF is also 100\%, while empty output leads to 0\% of the three scores. The better tracker is supposed to be higher for the three scores. Like IDF1, the SAIDF solves the problem of IDS and it concerns all segments instead.
    
    \vspace{-4mm} 
    \subsubsection{Comparison with Previous Measures.} As shown in Fig.~\ref{fig:IDS}, SAIDF score fixes the problems of IDS and IDF1 simultaneously. Fig.~\ref{fig:SEN} demonstrates the sensitivity to identity issues of different measures with a more general case. For a true trajectory with length $l_g$, we suppose a tracking result has three matched segments $\mathcal{S}_1$, $\mathcal{S}_2$, and $\mathcal{S}_3$ with different identities. The segments $\mathcal{S}_1$ and $\mathcal{S}_2$ has the same length $l=m\cdot l_g(0<m\le 0.5)$ and the segment $\mathcal{S}_3$ has length $l_c=l_g-2l$. One identity issue can be eliminated if we can combine the identities of two segments $\mathcal{S}_1$ and $\mathcal{S}_2$ into one identity. The improvements on MOTA, IDF1 and SAIDF of solving such identity issue are shown in Fig.~\ref{fig:SEN}. It shows that MOTA is not sensitive when $m$ is large, and IDF1 is not sensitive when $m$ is small, but for SAIDF, the improvement is always sensitive. 
    \vspace{-4mm} 
    \subsubsection{Computation.}  Calculating SAIDF is efficient, because it needs not to calculate the optimal matching relationships between true trajectories and hypotheses. The computational cost is linear with the number of target boxes in the results and ground-truth. Even in longer videos with more target boxes, the evaluation time of SAIDF increases slightly, while the evaluation time of IDF1 becomes unbearable. Please refer to supplementary material for details.  



    \begin{table}
    \scriptsize
    \centering
    \renewcommand{\arraystretch}{1.2}
    \begin{tabular}{|l|c|c|c|c|c|c|c|c|c|c|}
    \hline
    Processing &Method & \textbf{MOTA} $\uparrow$ & MOTP $\uparrow$ & \textbf{IDF1} $\uparrow$ & IDP $\uparrow$ & IDR $\uparrow$ & FP $\downarrow$ & FN $\downarrow$ & \textbf{IDS} $\downarrow$ \\
    \hline\hline
    
    \multirow{7}*{ \centering
        Online
    } & FAMNet\cite{Chu_2019_ICCV} & \textcolor[rgb]{0.00,0.00,1.00}{\textbf{52.0\%}} & \textcolor[rgb]{0.00,0.00,0.00}{76.5\%} & \textcolor[rgb]{0.00,0.00,0.00}{48.7\%} & \textcolor[rgb]{0.00,0.00,0.00}{66.7\%} & \textcolor[rgb]{0.00,0.00,0.00}{38.4\%} & \textcolor[rgb]{0.00,0.00,1.00}{\textbf{14138}} & 253616 & \textcolor[rgb]{0.00,0.00,0.00}{3072} \\

    ~ & STRN\_MOT17\cite{Xu_2019_ICCV} & 50.9\% & \textcolor[rgb]{0.00,0.00,0.00}{75.6\%} & \textcolor[rgb]{0.00,0.00,1.00}{\textbf{56.0\%}} & 74.4\% & \textcolor[rgb]{1.00,0.00,0.00}{\textbf{44.9\%}} & \textcolor[rgb]{0.00,0.00,0.00}{25295} & \textcolor[rgb]{0.00,0.00,1.00}{\textbf{249365}} & 2397 \\
    
    ~ & MOTDT17\cite{long2018real} & 50.9\% & \textcolor[rgb]{0.00,0.00,1.00}{\textbf{76.6\%}} & \textcolor[rgb]{0.00,0.00,0.00}{52.7\%} & \textcolor[rgb]{0.00,0.00,0.00}{70.4\%} & \textcolor[rgb]{0.00,0.00,0.00}{42.1\%} & \textcolor[rgb]{0.00,0.00,0.00}{24069} & 250768 & \textcolor[rgb]{0.00,0.00,0.00}{2474} \\
    
    ~ & MTDF17\cite{fu2019multi} & 49.6\% & 75.5\% & 45.2\% & 58.1\% & 37.0\% & 37124 & \textcolor[rgb]{1.00,0.00,0.00}{\textbf{241768}} & 5567 \\

    ~ & PHD\_GM\cite{sanchez2019predictor} & 48.8\% & \textcolor[rgb]{1.00,0.00,0.00}{\textbf{76.7\%}} & 43.2\% & 58.2\% & 34.3\% & 26260 & 257971 & 4407 \\

    ~ & DMAN\cite{zhu2018online} & 48.2\% & 75.7\% & 55.7\% & \textcolor[rgb]{0.00,0.00,1.00}{\textbf{75.9\%}} & 44.0\% & 26218 & 263608 & \textcolor[rgb]{0.00,0.00,1.00}{\textbf{2194}} \\
            
    ~ & Ours & \textcolor[rgb]{1.00,0.00,0.00}{\textbf{52.4\%}} & \textcolor[rgb]{1.00,0.00,0.00}{\textbf{76.7\%}} & \textcolor[rgb]{1.00,0.00,0.00}{\textbf{56.3\%}} & \textcolor[rgb]{1.00,0.00,0.00}{\textbf{76.9\%}} & \textcolor[rgb]{0.00,0.00,1.00}{\textbf{44.4\%}} & \textcolor[rgb]{1.00,0.00,0.00}{\textbf{14081}} & 252236 & \textcolor[rgb]{1.00,0.00,0.00}{\textbf{2166}}
    
    \cr\cline{2-10}
    
    ~ & Tracktor17\cite{Bergmann_2019_ICCV} & \textcolor[rgb]{0.00,0.00,0.00}{53.5\%} & \textbf{78.0}\% & \textcolor[rgb]{0.00,0.00,0.00}{52.3\%} & \textcolor[rgb]{0.00,0.00,0.00}{71.1\%} & \textcolor[rgb]{0.00,0.00,0.00}{41.4\%} & \textcolor[rgb]{0.00,0.00,0.00}{12201} & 248047 & \textcolor[rgb]{.00,0.00,0.00}{\textbf{2072}} \\
    
    ~ & Ours with FPN & \textcolor[rgb]{0.00,0.00,0.00}{\textbf{55.0\%}} & 77.9\% & \textcolor[rgb]{0.00,0.00,0.00}{\textbf{57.2\%}} & \textcolor[rgb]{0.00,0.00,0.00}{\textbf{76.8\%}} & \textcolor[rgb]{0.00,0.00,0.00}{\textbf{45.6\%}} & \textcolor[rgb]{0.00,0.00,0.00}{\textbf{11502}} & \textbf{240342} & \textcolor[rgb]{0.00,0.00,0.00}{2126} \\
    
    \hline

    \hline
    \end{tabular}
    \renewcommand{\arraystretch}{1}
     \vspace{3mm} 
    \caption{Comparison of the proposed MOT framework with other online processing state-of-the-art (SOTA) methods in MOT17 whose MOTA is larger than 48\%. \textcolor{red}{red} means the best result and \textcolor{blue}{blue} denotes the runner-up. $\uparrow$ means higher is better and $\downarrow$ means lower is better. We compare  with \textit{Tracktor17} separately by adding FPN as post-processing, \textbf{bold} for better result between ours(w. FPN) and \textit{Tracktor17}. }
    \label{tab:resol}
    \end{table}


    \section{Experiments}
 \vspace{-2mm}    

    \subsection{Implementation Details}
    \vspace{-1mm} 
    
    
    The network is trained in three stages. In the first stage, the appearance feature extractor and the backbone are trained on ReID datasets market-1501\cite{zheng2015scalable}. The backbone is pretrained on ImageNet\cite{ILSVRC15} and other parameters are randomly initialized. Afterwards, the mAP of the appearance feature extractor on market-1501 dataset is 62\%.   For the second stage, the switcher heat-map generator is trained on Youtube-BB\cite{real2017youtube} dataset but only pedestrians are considered. In the third stage, the whole network is trained on MOT16\cite{MOT16} dataset supervised by the losses mentioned in Sec.~\ref{sec:APPR}. MOT datasets are only used in the last stage and the first two stages are pretraining steps for the final decision-maker.

    In all stages of training we used SGD optimizer with momentum of 0.9. In the third stage, we applied initial learning rate of 0.1 and decay by 0.1 at epoch 6, 10 and 15. We implemented training for 20 epochs in total. It is vital in the third training stage to generate the training data using detection-based bounding boxes because the ground truth is much more precise than practical detections. Training on ground truth will lead to an over-fitting problem to easy cases, and thus the network will not perform well in real detection-based scene.

    \vspace{-4mm} 

    \subsection{Evaluation on MOT benchmarks}
    \vspace{-1mm} 
    \subsubsection{Datasets.}
    The proposed framework is evaluated with the MOT17 benchmark \cite{MOT16}, which shares the same training and test videos with MOT16 but offers different detection input. MOT17 has made the ground truth more accurate. Besides, it requires the algorithm to be evaluated under three different detectors simultaneously, making the results more convincing. The test video sequences include various complicated scenes and are still a great challenge. The benchmark uses CLEAR MOT Metric and IDF1 score for evaluation.

 \vspace{-4mm}    
    
    \subsubsection{Results.}
    Table~\ref{tab:resol} illustrates the results for state-of-the-art  (SOTA) online processing methods and our approach in MOT17 testing set.
     The results show that our framework has highest MOTA when compared to some popular methods. At the same time, our framework has the least IDS number and the highest IDF1 scores. In addition, our MOTA, IDS and IDF1 scores are in the leading positions for MOT17 for all the online processing algorithms considered. High IDF1 and low IDS also demonstrate that our method works well on solving identity issues. When comparing with the method \textit{Tracktor17}\cite{Bergmann_2019_ICCV}, we find that a FPN is used in the work. For fair comparison, we also add the same FPN from \textit{Tracktor17} as post-processing for our result. Results show that our work outperforms \textit{Tracktor17} under the FPN setting in terms of most measurements such as MOTA, IDF1 and so on. As for MOTP and IDS, our algorithm is almost comparable to \textit{Tracktor17}.

    Besides, the backbone we used is ResNet-18, while many other works like \cite{zhu2018online,long2018real} use larger networks such as ResNet-50 for appearance feature extraction. We achieve competitive results on multiple measures by a smaller network, which also indicates the effectiveness of switcher information. With the help of deeper networks and spatial-temporal techniques on feature extracting, we believe the switcher information may yield larger improvement in the future.
    
    \vspace{-2mm}
   
    \begin{table}
    \footnotesize
    \centering
    \begin{tabular}{|l|c|c|c|c|}
    \hline
    Method & \textbf{SAIDF} $\uparrow$ & \textbf{MOTA} $\uparrow$ & \textbf{IDF1} $\uparrow$   & \textbf{IDS} $\downarrow$ \\
    \hline\hline
    
     SST\cite{Sun2018DeepAN} & \textcolor[rgb]{0.00,0.00,0.00}{46.5\%} & \textcolor[rgb]{0.00,0.00,0.00}{45.0\%} & \textcolor[rgb]{0.00,0.00,0.00}{52.0\%} & 
    \textcolor[rgb]{0.00,0.00,0.00}{3555} \\
    
     MOTDT\cite{long2018real} & \textcolor[rgb]{0.00,0.00,0.00}{50.1\%} & \textcolor[rgb]{0.00,0.00,0.00}{51.0\%} & \textcolor[rgb]{0.00,0.00,0.00}{55.9\%} & 
    \textcolor[rgb]{0.00,0.00,0.00}{1429} \\

     deepmot\cite{xu2019deepmot} & \textcolor[rgb]{0.00,0.00,0.00}{51.3\%} & \textcolor[rgb]{0.00,0.00,0.00}{51.2\%} & \textcolor[rgb]{0.00,0.00,0.00}{56.6\%} & 
    \textcolor[rgb]{0.00,0.00,0.00}{1557} \\
    
     ours & \textcolor[rgb]{1.00,0.00,0.00}{\textbf{55.8\%}} & \textcolor[rgb]{1.00,0.00,0.00}{\textbf{55.2\%}} & \textcolor[rgb]{1.00,0.00,0.00}{\textbf{60.9\%}} & 
    \textcolor[rgb]{1.00,0.00,0.00}{\textbf{1064}} \\
    \hline
    \end{tabular}

     \vspace{3mm} 
    \caption{ Results on MOT17 training set comparing SAIDF, MOTA, IDF1 and IDS. \textcolor{red}{\textbf{red}} represents the best result. $\uparrow$ means higher is better and $\downarrow$ means lower is better.}
    \label{tab:resmk}
    \end{table} 
    
    
    \subsubsection{Results on SAIDF Measure.}
      Table \ref{tab:resmk} illustrates the results of some previous open-source online processing methods and our approach in MOT17 training set using the SAIDF measure that we propose in this work, together with the other previous measures such as MOTA, IDF1 and IDS. For the MOTA and IDF1 measures, our SAMOT method outperforms the other three open-source methods with large margin(55.2\% vs the SOTA of 51.2\% for MOTA; 60.9\% vs the SOTA of 56.6\% for IDF1). As for the new SAIDF, the proposed SAMOT method also yields a large margin (55.8\% vs 51.3\%).  Consistent improvement in terms of all these measures validate the effectiveness of the proposed approach. More importantly, it also indicates the effectiveness of the proposed SAIDF measure. In other words, the great reduction of identity issues is properly reflected with new SAIDF measure by relative improvement of 19.8\% (from 46.5\% of SST method to 55.8\% of our approach). While for IDF1, the same reduction leads to relative improvement of 17.1\%.
      

      \vspace{-4mm} 
    \subsection{Ablation Study and Discussion}
    \begin{figure}
    \centering
    \subfigure[]{
    \centering
    \label{fig:abla}
    \includegraphics[width=2.74in]{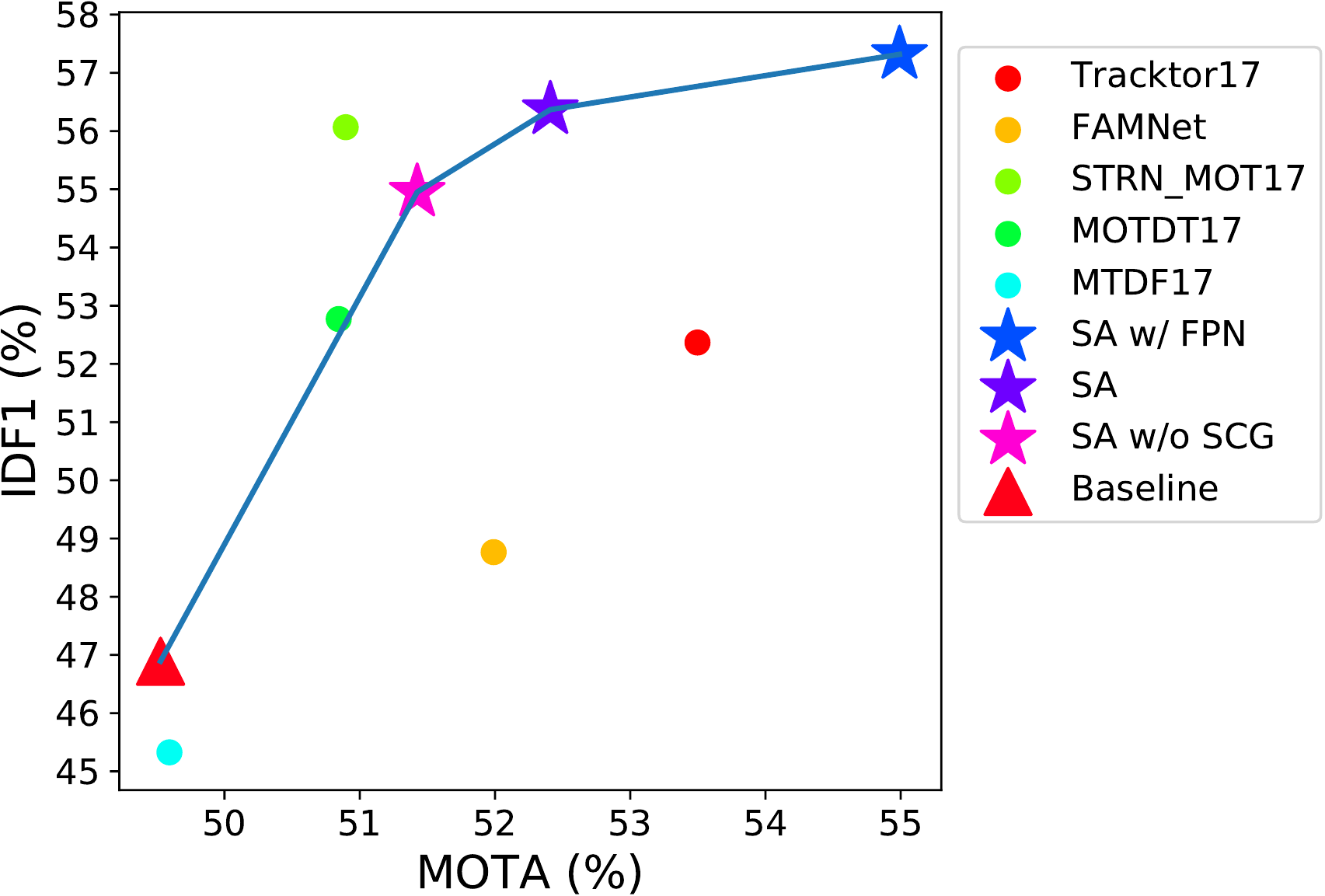}
    }
    \subfigure[]{
    \centering
    \label{fig:diss}
    \includegraphics[width=1.83in]{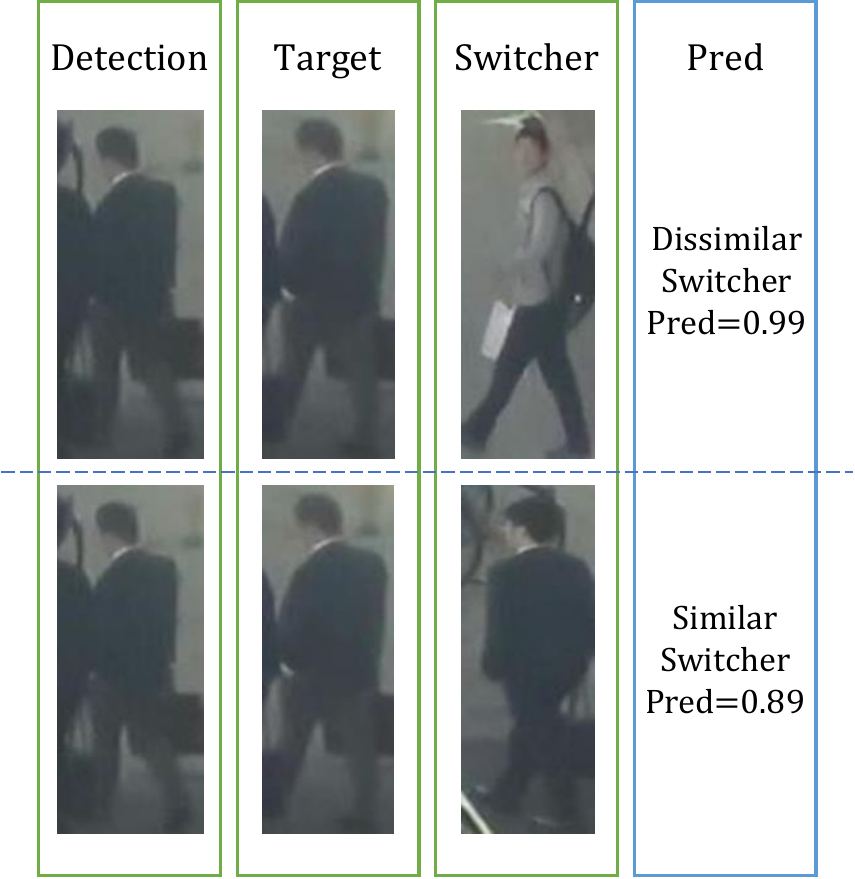}
    }
    \vspace{-3mm} 
       \caption{(a) Ablation study on our framework. SA w/o SCG in pink star stands for using SAA only. SA in purple star means using both SAA and SCG. SA w/ FPN represents using SAA and SCG with FPN as post-process. (b) Different prediction scores (Pred) obtained by the decision-maker on the same matching target-detection pair with different potential switchers.}
    
    \end{figure}  
    
    \subsubsection{How Different Components Impact the Tracking Performance?}
    Fig.~\ref{fig:abla} shows the impact of different components of the framework. We test on three detectors in MOT17 test dataset and the results are obtained from the average of the three detectors. The baseline method utilizes single target appearance association with binary classifier to refine detection score. As shown in Fig. \ref{fig:arch}, without the help of SCG, the performance decreases by 1.0\% MOTA and 1.4\% IDF1, and without the help of SAA (the baseline in Fig.~\ref{fig:abla}), the performance decreases by 1.9\% MOTA and 8.0\% IDF1. With the help of FPN, we further gain another 2.6\% MOTA and 0.9\% IDF1 improvement. With SAA, the edges in association is more discriminative so that more identity issues are solved and thus the IDF1 is 8.0\% higher. With SCG, more confusing spatial switchers are eliminated so that the IDF1 increases another 1.4\%.
    
    \vspace{-4mm} 
    \subsubsection{In What Cases Can the Switcher Information Help the Tracking?}
    With switcher information, the tracker is able to make a comparison between the target and its potential switcher, thus it has a higher accuracy when matching. In some cases that nearby targets have similar appearance, the decision-maker is trained to learn which one is better. In the cases that we could not find a similar potential switcher, the decision-maker learns to use a lower threshold for the appearance feature, which may be of low quality due to occlusions by background objects. As shown in Fig. \ref{fig:diss}, the decision-maker is trained to yield higher matching score if the potential switcher is not similar to the target.

\section{Conclusions}
\vspace{-2mm} 
In this paper, we have presented a novel switcher aware MOT framework that integrates SCG and SAA effectively together with a new MOT measure.
The SCG can filter out the false positives and inaccurate detections by analyzing the neighboring features in the same frame, while SAA provides more accurate data association with the aid of switcher heat map and appearance features. We also find that the previous measures such as IDS, MOTA and IDF1 are not sensitive to identity issues, so we design a new MOT measure, that is, SAIDF, to effectively reflect such related changes in the tracking results. Extensive experiments demonstrate that proposed SAMOT framework outperforms the SOTA MOT algorithms in challenging dataset MOT17. It is also shown that the new SAIDF measure may provide more insightful analysis about the identity issues. 

\clearpage

\setcounter{section}{0}
\begin{center}
\normalfont \huge
     Supplementary Materials
\end{center}

\renewcommand\thesection{A\arabic{section}}
\renewcommand\theequation{A\arabic{equation}}
\renewcommand\thefigure{A\arabic{figure}}
\renewcommand\thetable{A\arabic{table}}

\section{Comparison on True Videos}
\vspace{-20pt}
\begin{figure}
    \centering
    \subfigure[similar switcher: before the identity issue happened]{\includegraphics[width=2.3in]{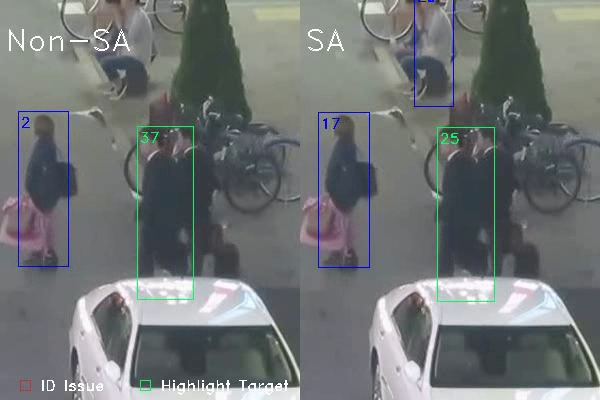}}
    \subfigure[similar switcher: after the identity issue happened]{\includegraphics[width=2.3in]{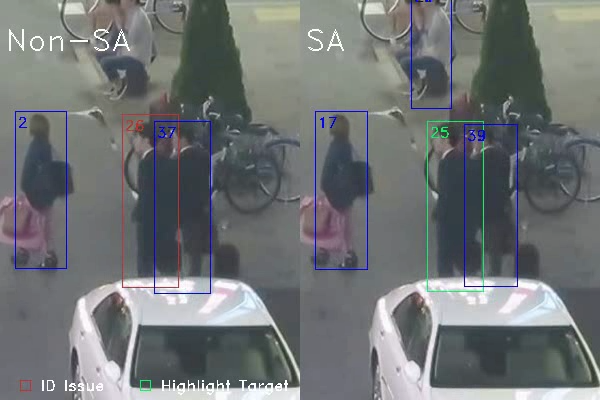}}
    \subfigure[dissimilar switcher: before the identity issue happened]{\includegraphics[width=2.3in]{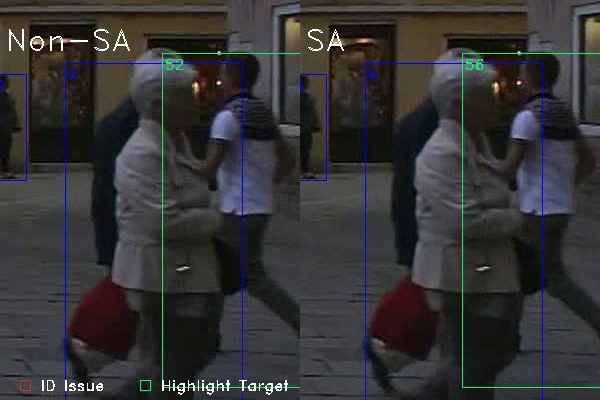}}
    \subfigure[dissimilar switcher: after the identity issue happened]{\includegraphics[width=2.3in]{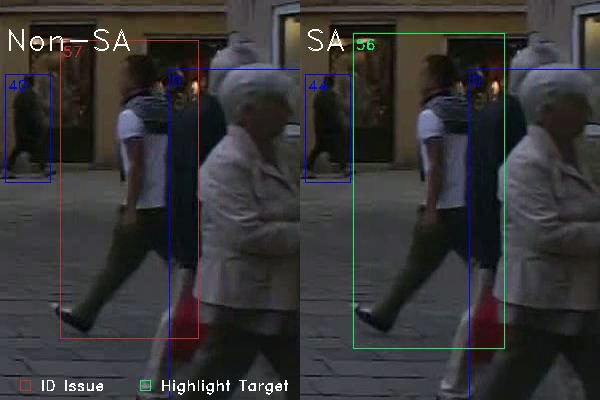}}
    \vspace{-3mm}
    \caption{Screen-shot of the demo video. Green box for highlighted target, red box for wrongly assigned target.}
    \label{fig:demo}
\end{figure}
\vspace{-4pt}
Video file named `demo.avi' shows the demo results of our method. As shown in Fig.~\ref{fig:demo}, it compares the `Non-SA' (baseline) with `SA' (baseline with our our switcher aware design). In the video, green boxes stand for highlighted targets and red boxes denote identity issues.
 These identity issues happening in Non-SA show that confusing switchers are one of the important error sources that cause identity issues. The proposed scheme can effectively overcome the identity issues in the video demonstration. 
\subsubsection{Similar Switchers}
In the case shown in 
Fig.~\ref{fig:demo} (a) and (b), the identity 37 is drifted to another target in Non-SA but the identity keeps unchanged in SA. In fact the identity 26 is a temporal switcher from previous frames, the inclusion of the potential switcher in our approach results in more reliable matching relationships. The decision-maker is trained to make fair judgement based on all appearance features and heat maps.
\subsubsection{Dissimilar Switchers}
In the case shown in Fig.~\ref{fig:demo} (c) and (d), the identity is changed from 52 to 57 for Non-SA but the identity is not changed for SA. This is an example where the similarity of the features for the highlighted target is low due to partial occlusion. Comparison to a dissimilar potential switcher allows a lower threshold to be used in our SA for matching the correct pair of targets in adjacent frames and thus the matching probability is increased relatively. It reflects another situation discussed in the ablation study of the main paper. \subsubsection{Failure Cases}
Although the inclusion of the switcher information solves many identity issues compared to non-switcher-aware, there are still some cases remain unsolved even we consider the potential switcher. This is probably because the quality of the appearance feature is extremely low, which may need more research in the future.

\section{Computational Complexity of SAIDF}
\vspace{-2mm}
Assume that the hypothesis output results of a tracker is $$\mathcal{\hat{H}}=\{\mathbf{h}_i=(h_i^{fr}, h_i^{id}, h_i^{pos})\}\\ | i=1,2,...,N_H\},$$ the ground truth set is $$\mathcal{\hat{G}}=\{\mathbf{g}_j=(g_j^{fr}, g_j^{id}, g_j^{pos})\}| j=1,2,...,N_G\},$$ where $h_i^{fr}, h_i^{id}, h_i^{pos}$ respectively stand for frame index, identity number, and position of the $i^{th}$ hypothesis output. The $j^{th}$ ground truth $\mathbf{g}_j$ also has similar meaning. 
Denote the identity index set of hypothesis output from a tracker as $ID_H=\{h_i^{id} | (h_i^{fr}, h_i^{id}, h_i^{pos}) \in \mathcal{\hat{H}}\}$ with size $N_{IH}$. Similarly, we have the identity index set of ground truth $ID_G=\{g_j^{id} | (g_j^{fr}, g_j^{id}, g_j^{pos}) \in \mathcal{\hat{G}}\}$ with size $N_{IG}$. Since the ground truth identity in different frames can be the same, i.e. $\exists g_j^{id} = g_k^{id}$ where $j\neq k$, we have $N_{IG} \leq N_G$. Similarly, $N_{IH}\leq N_H$. 
The matching relationships between the output and the ground truth (matched pairs) are denoted by $\mathcal{\hat{M}} = \{(u_k, v_k)|k=1,2,...,N_M\}$, where the $k^{th}$ matched pair $(u_k, v_k)$ means $\mathbf{h}_{u_k}\in  \mathcal{\hat{H}}$ and $\mathbf{g}_{v_k}\in \mathcal{\hat{G}}$ are matched. 
It is obvious that $N_M\leq N_H+N_G$. 
To compute SAIDF, the following two steps should be implemented:  First, in order to calculate SAIDR and SAIDP, we compute all $\lambda(o, q)$ or $\lambda(q, o)$ in Eq.(13) of the main paper where $o\in \mathcal{O},q\in \mathcal{Q}$. After optimization, the computational complexity for this step is $O(N_M)$. Second, we will enumerate all hypothesis-ground-truth identity pairs. The computational complexity for this step is $O(N_{\mathcal{Q}}*N_{\mathcal{O}})$, where $N_{\mathcal{Q}}$ and $N_{\mathcal{O}}$ are respectively the size of $\mathcal{Q}$ and $\mathcal{O}$. Because each identity number is assigned to a trajectory, we have $N_\mathcal{Q}=N_{IH}$ and $N_\mathcal{O}=N_{IG}$. Therefore, the complexity for the second step is $O(N_{IH}*N_{IG})$. The overall complexity of  
SAIDF is: \begin{equation}O(N_M)+O(N_{IH}*N_{IG}).\end{equation} 
For IDF1, it needs to solve an extra optimal matching problem.  The complexity of solving it using Hungary Algorithm is $O(N_{IH} * N_{IG}*(N_{IH}+N_{IG}))$. So the overall complexity for IDF1 is
\begin{equation}\begin{split} &O(N_M) +O(N_{IH}*N_{IG})\\ + &O(N_{IH} * N_{IG}*(N_{IH}+N_{IG})).\end{split}\end{equation}

   \begin{figure}[t]
      \centering
      \includegraphics[width=2.8in]{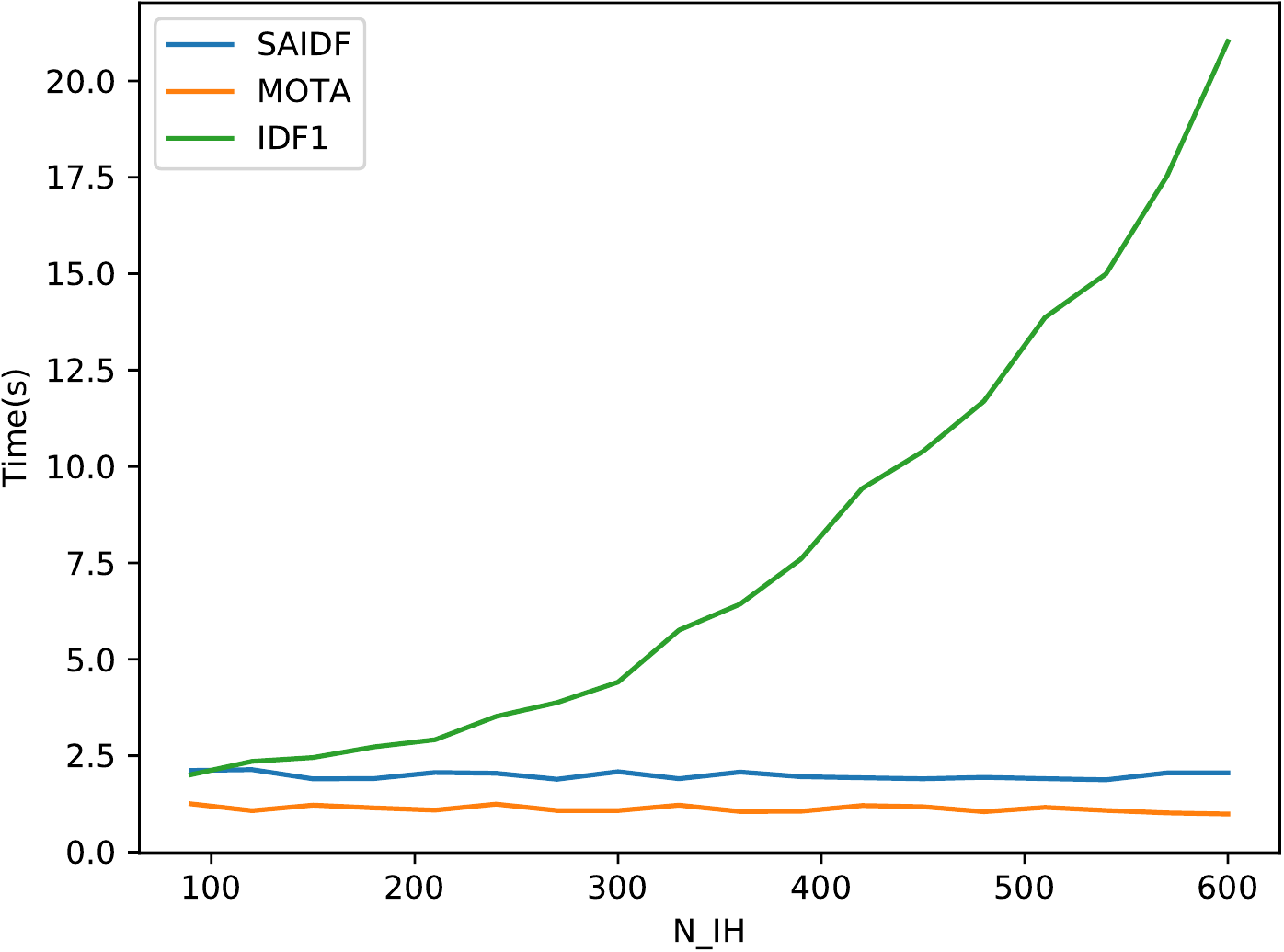}
      \vspace{-3mm}
      \caption{Time cost with different $N_{IH}$ for one video (implemented in python, running on i7-8750H@2.20GHZ with single core) }
      \label{fig:my_fig}
   \end{figure}
Although $N_{IG}$ is not affected by the tracker and is far less than $N_G$, $N_{IH}$ is always affected by the tracker and in the worst case it can be up to $N_H$.
If the tracker performs good, the $N_{IH}$ will be usually small, then computing IDF1 will not be so computationally complicated. 
However, if the tracker is not so good, the $N_{IH}$ will be large, then it may take a long time to compute IDF1. Besides, if the video length is long and the $N_{IH}$ increases, the time cost of calculating IDF1 will increase non-linearly. Fig.\ref{fig:my_fig} shows the simulation results about how $N_{IH}$ affects the computing time of SAIDF, IDF1 and MOTA. SAIDF requires much less computational time than IDF1 when the outputs are not very good. 
In the simulation (implemented in python), we assume $N_{IG}=30$ and the length of the video sequence is $200$. $N_{IH}$ hypothesis identities are random generated but the total number of bounding boxes in the output remains unchanged, i.e., $N_H$ keeps the same. 

In the research community, longer video with larger $N_{IG}$ is gaining more attention. For example, in MOT19, the maximum $N_{IG}=1240$ (CVPR19-05), $N_{IH}$ can be more than 2500. \emph{When the dataset videos become longer and have higher density or have many videos to combine (e.g. MTMC), IDF1 requires much more time  to compute when compared with MOTA and our proposed SAIDF. }

\section{Scenes for Best Use of SAIDF}
In our evaluation of several methods with MOT17 dataset, SAIDF has a similar trend with IDF1 (Table \ref{tab:resmk}). Actually, it depends on the dataset involved. We found that for a tracking dataset if hard cases are imbalanced along time, IDF1 is not so sensitive to the ID-mistakes. In MOT17, hard case imbalance is not obvious, so SAIDF is highly correlated with IDF1, as shown in Table 2. We construct an imbalanced scene based on MOT17 segments by accelerating the fps in the second half of the segment to make it harder to track for the second half. In such setting, a better method(SAIDF 54.4, IDF 52.5) and a worse method(SAIDF 51.7, IDF 52.0) are tested and the improvement of SAIDF(2.7) is about 5 times as that of IDF1(0.5), showing SAIDF is better in representing the improvement. But in a balanced scene, this better tracker is 8.6 and 8.0 higher than the worse one in terms of SAIDF and IDF1, respectively. In many practical scenes, the degree of hard case imbalance is high(e.g., waiting for traffic lights, targets first stay and then move), where SAIDF is better. Besides, SAIDF runs faster than IDF1, so in crowded and long videos, SAIDF is better.

\section{SCG: Optimal Subgraph Solving and Complexity Analysis}
\vspace{-2mm}
    On a rough thought, conflict graph seems to be complicated. Actually, 
    the graph built in Section 3.1 in the main paper has sparse edges and is composed of many connected components. A connected component is a subgraph in which there is a path between any pair of nodes in the subgraph. The optimal subgraph of the conflict graph is the union of optimal subgraphs in the connected components. Therefore, we first enumerate all possible subgraphs in each connected component of the conflict graph and find out the optimal subgraphs of each connected component. Then the overall optimal subgraph is the union of such subgraphs.
    Due to low density of detection results and sparse edge connections, the average detection number in each connected component is expected to be a constant not larger than $C$ ($C\le 8$ in MOT16, MOT17). Therefore, the time complexity of enumeration to deal with $N$ detections is $O(\frac{N}{C}\cdot 2^C)$, when $C\leq 8$, $O(\frac{N}{C}\cdot 2^C)=O(32N)\approx O(N)$.
    
    In MOT17 dataset, the time cost of SCG algorithm is on par with the time cost of Hungarian Algorithm, which is far less than the time cost of deep learning model forwarding. Even in denser pedestrian tracking scenes, due to sparse connections, the time cost of SCG is still acceptable.

\section{SAA: Backbone Details}
\vspace{-2mm}
 In this work, we use a modified ResNet-18 \cite{he2016deep} as the weight-sharing backbone of the appearance feature extractor, and the switcher
 heat-map generator. The main differences from the original design are that we remove the strides in 
       block3 and block4 (namely $conv{3\_x}$ and $conv{4\_x}$). In order to keep the perception field unchanged, we add dilation\cite{yu2015multi} in these two layers, which are set to dilation 2 for block3 and 4 for block4.

\section{Details for MOT16/17 benchmarks}
\vspace{-2mm}
Table \ref{tab:res17} shows the results for the whole evaluation metrics on the MOT17 test set.
Table \ref{tab:res16} illustrates the main metrics of the MOT16 test set compared to several previous methods. Both in MOT17 and MOT16, our method outperforms most previous online processing methods in most measures.

This result on MOT16 is \textbf{NOT} an extra experiment result but the result of a subset of MOT17 that we have not modified.
 \begin{table}[t]
    \tiny
    \centering
\begin{tabular}{@{  }c@{ }c@{ }c@{ }c@{ }c@{ }c@{ }c@{ }c@{ }c@{ }c@{ }c@{ }c@{ }c@{ }c@{ }c@{ }c@{ }c@{ }c@{ }}
\\
  & IDF1 & IDP & IDR & Rcll & Prcn & FAR & GT & MT & PT & ML & FP & FN & IDs & FM & M/A & M/P & M/AL \\
SA \\
\hline
       overall &     56.3 &     76.9 &     44.4 &     55.3 &     95.7 &     0.79 &   2355 &    383 &   1098 &    874 &  14081 & 252236 &   2166 &   8019 &     52.4 &     76.7 &     52.8 \\
  M/17-01-DPM &     43.1 &     71.3 &     30.8 &     41.2 &     95.4 &     0.28 &     24 &      4 &     11 &      9 &    128 &   3790 &     32 &    118 &     38.8 &     71.7 &     39.2 \\
  M/17-03-DPM &     61.3 &     80.6 &     49.5 &     60.0 &     97.7 &     0.97 &    148 &     38 &     85 &     25 &   1459 &  41837 &    219 &   1032 &     58.4 &     75.9 &     58.6 \\
  M/17-06-DPM &     53.0 &     75.3 &     40.9 &     49.0 &     90.3 &     0.52 &    222 &     23 &    107 &     92 &    619 &   6005 &     77 &    355 &     43.1 &     70.1 &     43.8 \\
  M/17-07-DPM &     46.4 &     70.5 &     34.6 &     45.7 &     93.1 &     1.14 &     60 &      6 &     32 &     22 &    569 &   9173 &    103 &    366 &     41.7 &     73.0 &     42.3 \\
  M/17-08-DPM &     32.5 &     68.7 &     21.2 &     29.4 &     95.0 &     0.52 &     76 &      6 &     32 &     38 &    325 &  14921 &     87 &    205 &     27.4 &     78.2 &     27.8 \\
  M/17-12-DPM &     53.5 &     79.3 &     40.3 &     47.0 &     92.5 &     0.37 &     91 &     16 &     33 &     42 &    332 &   4592 &     23 &    106 &     42.9 &     76.0 &     43.2 \\
  M/17-14-DPM &     41.1 &     82.5 &     27.4 &     30.8 &     93.0 &     0.57 &    164 &      9 &     63 &     92 &    428 &  12783 &     78 &    383 &     28.1 &     74.3 &     28.5 \\
M/17-01-FRCNN &     48.3 &     67.8 &     37.5 &     49.1 &     88.8 &     0.89 &     24 &      5 &     11 &      8 &    399 &   3284 &     12 &     72 &     42.7 &     76.7 &     42.9 \\
M/17-03-FRCNN &     62.4 &     80.3 &     51.1 &     62.5 &     98.4 &     0.73 &    148 &     44 &     80 &     24 &   1092 &  39212 &    186 &    640 &     61.3 &     78.1 &     61.5 \\
M/17-06-FRCNN &     54.9 &     71.4 &     44.5 &     55.8 &     89.4 &     0.65 &    222 &     36 &    112 &     74 &    780 &   5213 &     97 &    337 &     48.3 &     73.5 &     49.1 \\
M/17-07-FRCNN &     39.9 &     64.1 &     28.9 &     40.6 &     90.0 &     1.53 &     60 &      4 &     32 &     24 &    764 &  10029 &    123 &    385 &     35.4 &     75.2 &     36.1 \\
M/17-08-FRCNN &     33.0 &     71.3 &     21.5 &     27.8 &     92.1 &      0.8 &     76 &      7 &     29 &     40 &    502 &  15260 &     59 &    162 &     25.1 &     79.8 &     25.4 \\
M/17-12-FRCNN &     42.0 &     76.0 &     29.0 &     34.8 &     91.2 &     0.32 &     91 &      7 &     31 &     53 &    290 &   5655 &     36 &    168 &     31.0 &     77.0 &     31.4 \\
M/17-14-FRCNN &     38.8 &     69.6 &     26.9 &     33.3 &     86.1 &     1.32 &    164 &      8 &     74 &     82 &    991 &  12327 &    167 &    539 &     27.0 &     72.2 &     27.9 \\
  M/17-01-SDP &     54.8 &     71.0 &     44.7 &     56.6 &     90.0 &      0.9 &     24 &      8 &     10 &      6 &    406 &   2798 &     29 &    158 &     49.9 &     75.2 &     50.3 \\
  M/17-03-SDP &     69.8 &     79.0 &     62.6 &     78.0 &     98.4 &     0.87 &    148 &     76 &     58 &     14 &   1311 &  23075 &    246 &   1021 &     76.5 &     78.3 &     76.7 \\
  M/17-06-SDP &     55.2 &     70.6 &     45.4 &     57.6 &     89.6 &     0.66 &    222 &     41 &    110 &     71 &    785 &   5000 &     90 &    401 &     50.1 &     71.8 &     50.9 \\
  M/17-07-SDP &     48.1 &     66.5 &     37.6 &     52.2 &     92.4 &     1.45 &     60 &      9 &     35 &     16 &    727 &   8072 &    121 &    431 &     47.2 &     75.8 &     47.9 \\
  M/17-08-SDP &     36.1 &     68.7 &     24.5 &     33.8 &     94.8 &     0.62 &     76 &     10 &     31 &     35 &    390 &  13994 &    149 &    353 &     31.2 &     79.3 &     31.9 \\
  M/17-12-SDP &     50.2 &     78.5 &     36.9 &     42.9 &     91.4 &     0.39 &     91 &     13 &     32 &     46 &    349 &   4946 &     36 &    178 &     38.5 &     77.4 &     38.9 \\
  M/17-14-SDP &     47.2 &     68.9 &     36.0 &     44.4 &     85.1 &     1.91 &    164 &     13 &     90 &     61 &   1435 &  10270 &    196 &    609 &     35.6 &     73.6 &     36.7 \\
\hline
\\
SA+FPN \\
\hline
       overall &     57.2 &     76.8 &     45.6 &     57.4 &     96.6 &     0.65 &   2355 &    432 &   1051 &    872 &  11502 & 240342 &   2126 &   6194 &     55.0 &     77.9 &     55.4 \\
  M/17-01-DPM &     40.8 &     70.9 &     28.7 &     38.6 &     95.4 &     0.26 &     24 &      5 &      7 &     12 &    119 &   3958 &     31 &     76 &     36.3 &     75.6 &     36.8 \\
  M/17-03-DPM &     63.9 &     78.8 &     53.7 &     66.9 &     98.1 &     0.88 &    148 &     53 &     76 &     19 &   1326 &  34639 &    234 &    409 &     65.4 &     78.6 &     65.6 \\
  M/17-06-DPM &     56.7 &     80.5 &     43.8 &     52.9 &     97.2 &     0.15 &    222 &     36 &     96 &     90 &    178 &   5549 &     76 &    195 &     50.8 &     80.3 &     51.4 \\
  M/17-07-DPM &     44.7 &     71.1 &     32.6 &     43.5 &     94.8 &      0.8 &     60 &      6 &     29 &     25 &    401 &   9551 &    100 &    286 &     40.5 &     75.7 &     41.1 \\
  M/17-08-DPM &     32.1 &     70.3 &     20.8 &     28.4 &     96.1 &     0.39 &     76 &      7 &     31 &     38 &    243 &  15133 &     67 &    128 &     26.9 &     81.3 &     27.2 \\
  M/17-12-DPM &     54.4 &     83.3 &     40.4 &     47.1 &     97.2 &     0.13 &     91 &     16 &     32 &     43 &    117 &   4584 &     24 &     64 &     45.5 &     80.9 &     45.7 \\
  M/17-14-DPM &     39.2 &     81.1 &     25.8 &     29.9 &     93.7 &      0.5 &    164 &      8 &     64 &     92 &    373 &  12965 &     80 &    348 &     27.4 &     72.2 &     27.8 \\
M/17-01-FRCNN &     46.6 &     76.1 &     33.6 &     41.9 &     94.9 &     0.32 &     24 &      5 &      9 &     10 &    145 &   3750 &     10 &     92 &     39.5 &     75.1 &     39.6 \\
M/17-03-FRCNN &     63.8 &     78.6 &     53.7 &     67.1 &     98.2 &     0.83 &    148 &     49 &     78 &     21 &   1251 &  34446 &    210 &    375 &     65.7 &     78.5 &     65.9 \\
M/17-06-FRCNN &     57.6 &     77.0 &     46.0 &     57.2 &     95.7 &     0.25 &    222 &     42 &    107 &     73 &    301 &   5049 &     85 &    247 &     53.9 &     79.7 &     54.6 \\
M/17-07-FRCNN &     40.9 &     66.7 &     29.5 &     41.6 &     93.9 &     0.91 &     60 &      6 &     28 &     26 &    456 &   9870 &    111 &    246 &     38.2 &     76.1 &     38.9 \\
M/17-08-FRCNN &     33.4 &     73.8 &     21.6 &     28.3 &     96.6 &     0.33 &     76 &      8 &     30 &     38 &    209 &  15156 &     55 &    107 &     27.0 &     80.9 &     27.3 \\
M/17-12-FRCNN &     48.6 &     78.6 &     35.1 &     43.2 &     96.8 &     0.14 &     91 &     13 &     28 &     50 &    125 &   4921 &     29 &     36 &     41.4 &     81.0 &     41.8 \\
M/17-14-FRCNN &     37.3 &     69.9 &     25.4 &     31.7 &     87.3 &     1.14 &    164 &      8 &     70 &     86 &    855 &  12616 &    147 &    434 &     26.3 &     71.7 &     27.1 \\
  M/17-01-SDP &     54.8 &     71.0 &     44.7 &     56.6 &     90.0 &      0.9 &     24 &      8 &     10 &      6 &    406 &   2798 &     29 &    158 &     49.9 &     75.2 &     50.3 \\
  M/17-03-SDP &     69.8 &     79.0 &     62.6 &     78.0 &     98.4 &     0.87 &    148 &     76 &     58 &     14 &   1311 &  23075 &    246 &   1021 &     76.5 &     78.3 &     76.7 \\
  M/17-06-SDP &     55.2 &     70.6 &     45.4 &     57.6 &     89.6 &     0.66 &    222 &     41 &    110 &     71 &    785 &   5000 &     90 &    401 &     50.1 &     71.8 &     50.9 \\
  M/17-07-SDP &     48.1 &     66.5 &     37.6 &     52.2 &     92.4 &     1.45 &     60 &      9 &     35 &     16 &    727 &   8072 &    121 &    431 &     47.2 &     75.8 &     47.9 \\
  M/17-08-SDP &     36.1 &     68.7 &     24.5 &     33.8 &     94.8 &     0.62 &     76 &     10 &     31 &     35 &    390 &  13994 &    149 &    353 &     31.2 &     79.3 &     31.9 \\
  M/17-12-SDP &     50.2 &     78.5 &     36.9 &     42.9 &     91.4 &     0.39 &     91 &     13 &     32 &     46 &    349 &   4946 &     36 &    178 &     38.5 &     77.4 &     38.9 \\
  M/17-14-SDP &     47.2 &     68.9 &     36.0 &     44.4 &     85.1 &     1.91 &    164 &     13 &     90 &     61 &   1435 &  10270 &    196 &    609 &     35.6 &     73.6 &     36.7 \\
\hline
    \end{tabular}
    \vspace{2mm}
    \caption{Details in MOT17 benchmarks. ``M/17" for ``MOT17", ``M/A" for ``MOTA", ``M/P" for ``MOTP", ``M/AL" for ``MOTAL".}
    \label{tab:res17}
\end{table}

\begin{table}
\scriptsize
\centering
\begin{tabular}{|l|c|c|c|c|c|}
\hline

Method & Ours & KCF~\cite{chu2019online} & STRN~\cite{Xu_2019_ICCV} &  MOTDT~\cite{long2018real} & AMIR~\cite{sadeghian2017tracking} \\
\hline\hline

MOTA & 53.8\% & 48.8\% & 48.5\% & 47.6\% & 47.2\% \\\hline
IDS & 606 & 906 & 747 & 792 & 774 \\\hline
IDF1 & 56.7\% & 47.2\% & 53.9\%  & 50.9\% & 46.3\% \\\hline

\end{tabular}
\vspace{2mm}
\caption{Performance on MOT16 with DPM detection results.}\label{tab:res16}

\end{table}

\clearpage

%
%
\bibliographystyle{splncs04}
\bibliography{egbib}
\end{document}